\documentclass[letterpaper, 10 pt, conference]{ieeeconf}  

\IEEEoverridecommandlockouts                              

\usepackage{graphicx}
\usepackage{caption}
\usepackage{booktabs}
\usepackage{multirow}
\usepackage{amsmath}
\usepackage{amssymb}
\usepackage{bm}
\usepackage{makecell}
\usepackage{subcaption}
\usepackage{wrapfig}

\usepackage{xcolor}
\usepackage{hyperref}
\usepackage{cleveref}

\hypersetup{
    colorlinks=true,
    linkcolor=blue,
    urlcolor=magenta 
}

\newcommand{\SE}{\mathrm{SE}(3)}

\renewcommand{\vec}[1]{\bm{#1}}
\newcommand{\mat}[1]{\mathbf{#1}}

\title{\LARGE \bf
HRDexDB: A 4D Dexterous Grasping Dataset Across\\ Human and Multiple Robot Embodiments
}

\author{
    \authorblockN{
        Jongbin Lim\textsuperscript{1,*}, 
        Taeyun Ha\textsuperscript{1,*}, 
        Seongho Cha\textsuperscript{1},
        Kanghyeon Cho\textsuperscript{1},
        Mingi Choi\textsuperscript{1}, 
        Subin Jeon\textsuperscript{1},
        Jisoo Kim\textsuperscript{1}, \\
        Byungjun Kim\textsuperscript{1}, and 
        Hanbyul Joo\textsuperscript{1,2}
    }
    \vspace{2mm} 
    \authorblockA{
        \textsuperscript{1}Seoul National University, \textsuperscript{2}RLWRLD 
        \vspace{1.5mm} \\ 
        \href{https://snuvclab.github.io/HRDexDB/}{\color{violet}\textbf{https://snuvclab.github.io/HRDexDB/}}
    }
}

\newcommand{\dataset}{HRDexDB}

\begin{document}

\maketitle
\thispagestyle{empty}
\pagestyle{empty}

\begin{abstract}
    We present \textbf{HRDexDB}, a real-world 4D dexterous grasping dataset capturing 3D hand-object interaction trajectories over time across five embodiments.
    The dataset comprises 3.2K trials over 100 diverse objects. Using a synchronized multi-camera system and an integrated reconstruction pipeline, HRDexDB provides multi-view and egocentric RGB observations, 3D hand geometry, robot states, and object 6D pose trajectories, together with success/failure annotations. 
    Human and robotic hands interact with shared objects, enabling the study of embodiment-dependent grasp strategies and contact patterns. 
    We demonstrate the dataset's utility through human-to-robot contact map transfer, visual robot-object contact estimation, and retrieval-assisted grasping.
    Together, these results establish HRDexDB as a resource for studying and learning dexterous interactions across human and robotic embodiments.
\end{abstract}


\begin{figure*}[t]
    \centering
    \includegraphics[width=\textwidth]
        {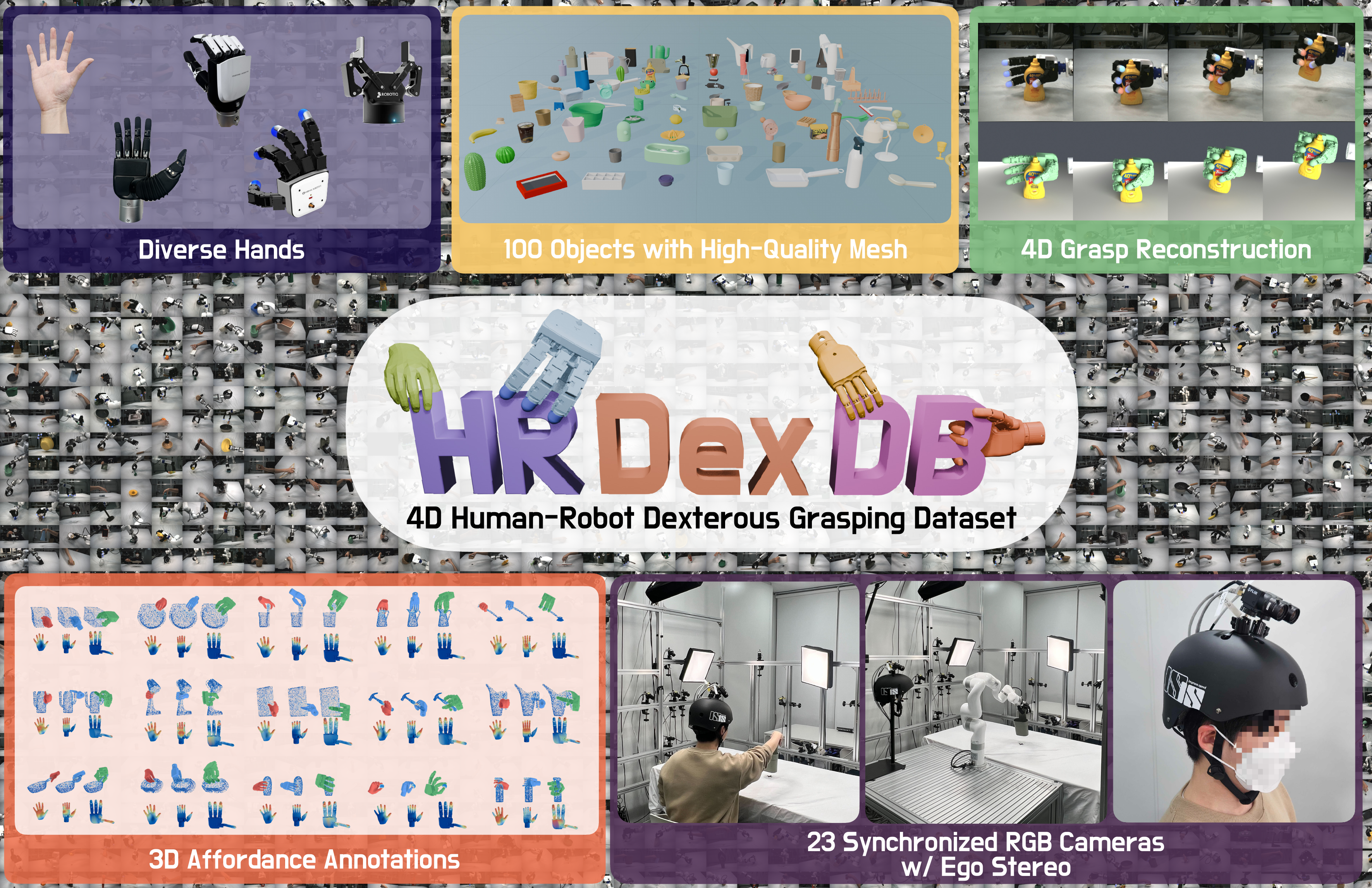}
    \caption{
        \textbf{Overview of HRDexDB.}
        HRDexDB contains 3.2K human and robotic dexterous grasping episodes
        across 100 objects and multiple hand embodiments. Using a synchronized
        multi-view capture system, we record human demonstrations
        and robotic executions with reconstructed 3D hand and robot trajectories,
        object 6D poses, egocentric observations, and success/failure
        annotations.
    }
    \label{fig:teaser}
    \vspace{-12pt}
\end{figure*}
\section{Introduction}
\label{sec:intro}

Enabling robots to achieve human-level dexterity is one of the central goals in robotics. Dexterous hands offer diverse ways to establish and maintain contact with objects, but their capabilities depend on morphology, kinematics, and actuation. Even when grasping the same object, human and robotic hands may adopt different finger configurations and contact patterns to achieve a stable grasp. Understanding these variations is important for developing grasping methods that accommodate different hand designs and for learning from demonstrations collected across embodiments. This motivates studying how human and robotic hands interact with a common set of objects in the real world. 

Existing datasets provide substantial resources for studying individual aspects of dexterous manipulation, but the current data landscape remains fragmented across embodiments and annotation types. 
Human-centric datasets provide either isolated hand motion without objects~\cite{moon2020interhand26mdatasetbaseline3d}, human hand-object interactions~\cite{garcia2018first, brahmbhatt2019contactdb, zimmermann2019freihand, Brahmbhatt_2020_ECCV, chao2021dexycb, liu2022hoi4d, fan2023arctic, zhan2024oakink2, wang2025hocap,citahampali2020honnotate}, or large-scale egocentric RGB videos without 3D information~\cite{grauman2022ego4dworld3000hours, Damen2021PAMI}. Robotic hand datasets \cite{bu2025agibot, o2024open, wu2025robocoin, khazatsky2024droid, wu2025robomind}, in contrast, focus on the robot side and often leave object motion only partially tracked. 
A few recent approaches attempt to connect demonstrations across embodiments~\cite{fang2024rh20t, tao2025dexwild, xie2026human2robot}, yet the combination of multiple dexterous hand designs, shared object coverage, and detailed multimodal observations remains limited. These gaps make it difficult to examine embodiment-dependent grasping behavior.

To address these limitations, we introduce \textbf{HRDexDB}, a real-world 4D dexterous grasping dataset containing 3.2K trials across five embodiments, comprising a human hand and four robotic hands, over a shared collection of 100 objects. 
HRDexDB captures the temporal evolution of 3D hand-object interactions through synchronized multi-view recording and reconstruction. 
The dataset provides reconstructed hand geometry and object 6D trajectories
in a unified world coordinate system, together with success/failure annotations.
HRDexDB is the first dataset to provide human and multiple robotic hands interacting with shared objects using a markerless multi-view RGB system.

To relate grasping behavior across embodiments, we associate human and robot trials through a reference-based
collection protocol: a teleoperator observes a human grasp and
reproduces its strategy while adapting it to the target robot's
morphology and kinematic constraints. These associations
represent a shared grasping intent without assuming identical
contacts or frame-level temporal alignment.


We demonstrate the utility of HRDexDB through three
downstream tasks spanning human-to-robot contact map transfer, visual
contact perception, and retrieval-assisted grasp synthesis. First, we study
\emph{human-to-robot contact map transfer}, where a model
predicts robot-specific contact patterns from a human grasp
while accounting for differences in hand morphology. The
transferred contacts improve grasp success over directly using
human contacts in both simulation and real-world experiments.
Second, we introduce \emph{visual robot-object contact
estimation}, using reconstructed robot geometry and tracked
object poses to derive region-level contact labels for
single-view RGB images, including regions occluded from the
camera.
Third, we study \emph{retrieval-assisted grasping}, which uses
local object geometry to retrieve robot grasps from our dataset and
adapts them to unseen objects through geometric alignment
and contact refinement.
Together, these tasks demonstrate how HRDexDB's demonstration-execution associations, geometric annotations, and recorded robot grasps support downstream robotics applications.

In summary, our contributions are threefold: (1) We introduce HRDexDB, a real-world dataset of human and robotic dexterous grasping over shared objects, with multimodal observations and detailed 3D annotations. 
(2) We present a unified pipeline for synchronized multi-view
capture and 3D reconstruction of human and robot
hand-object interactions.
(3) We establish downstream benchmarks on HRDexDB, including human-to-robot contact map transfer, RGB to robot-object binary contact estimation, and retrieval-assisted grasping.
At the time of submission, HRDexDB includes 100 captured objects, with ongoing expansion toward 1,000 objects.
The full dataset and benchmark protocols will be publicly released to facilitate future research in dexterous manipulation.

\section{Related Work}
\label{sec:related_work}

\begin{table*}[t]
  \centering
  \caption{Comparison of Human-Object Interaction (HOI) and Robotics Datasets with HRDexDB}
  \label{tab:dataset_comparison}
  \resizebox{\textwidth}{!}{ 
  \begin{tabular}{lcccccccccccc}
    \toprule
    \textbf{Dataset} & \textbf{Type} & \textbf{\#Emb.} &
    \textbf{Dex Robot Hand} & \textbf{Modality} & \textbf{Views} &
    \textbf{Objs} & \textbf{Resolution} & \textbf{Seqs} &
    \textbf{Frames} & \textbf{Markerless} & \textbf{3D Hand} &
    \textbf{Obj 6D} \\
    \midrule

    FPHA\cite{garcia2018first}
    & HOI & 1 & -- & RGB-D & 1 & 26 & $1920\times1080$
    & 1.2K & 105K & $\times$ & \checkmark & \checkmark \\

    ContactDB\cite{brahmbhatt2019contactdb}
    & HOI & 1 & -- & RGB-D & 9 & 50 & $1920\times1080$
    & 3.7K & 375K & \checkmark & $\times$ & \checkmark \\

    FreiHAND\cite{zimmermann2019freihand}
    & HOI & 1 & -- & RGB & 8 & 25 & $1280\times1024$
    & -- & 37K & \checkmark & \checkmark & $\times$ \\

    HO-3D\cite{citahampali2020honnotate}
    & HOI & 1 & -- & RGB-D & 5 & 10 & $640\times480$
    & 68 & 103K & \checkmark & \checkmark & \checkmark \\

    GRAB\cite{taheri2020grab}
    & HOI & 1 & -- & MoCap & -- & 51 & --
    & 1.3K & 1.6M & $\times$ & \checkmark & \checkmark \\

    ContactPose\cite{Brahmbhatt_2020_ECCV}
    & HOI & 1 & -- & RGB-D & 3 & 25 & $960\times540$
    & 2.3K & 2.9M & $\times$ & \checkmark & \checkmark \\

    DexYCB\cite{chao2021dexycb}
    & HOI & 1 & -- & RGB-D & 8 & 20 & $640\times480$
    & 1K & 582K & \checkmark & \checkmark & \checkmark \\

    HOI4D\cite{liu2022hoi4d}
    & HOI & 1 & -- & RGB-D & 1 & 800 & $1280\times720$
    & 4K & 2.4M & \checkmark & \checkmark & \checkmark \\

    ARCTIC\cite{fan2023arctic}
    & HOI & 1 & -- & RGB & 9 & 11 & $2800\times2000$
    & 339 & 2.1M & $\times$ & \checkmark & \checkmark \\

    TACO\cite{liu2024taco}
    & HOI & 1 & -- & RGB/RGB-D & 13 & 196 & $4096\times3000$
    & 2.5K & 5.2M & $\times$ & \checkmark & \checkmark \\

    OakInk2\cite{zhan2024oakink2}
    & HOI & 1 & -- & RGB-D & 16 & 75 & $840\times480$
    & 627 & 1.34M & $\times$ & \checkmark & \checkmark \\

    Contact4D\cite{songcontact4d}
    & HOI & 1 & -- & RGB & 19 & N.A. & $3840\times2160$
    & 375 & 2M & $\times$ & \checkmark & \checkmark \\

    HOT3D\cite{banerjee2025hot3d}
    & HOI & 1 & -- & RGB & 2 & 33 & $1408\times1408$
    & 425 & 3.7M & $\times$ & \checkmark & \checkmark \\

    GigaHands\cite{fu2025gigahands}
    & HOI & 1 & -- & RGB & 51 & 417 & $1280\times720$
    & 14K & 183M & \checkmark & \checkmark & \checkmark \\

    HO-Cap\cite{wang2025hocap}
    & HOI & 1 & -- & RGB-D & 9 & 64 & $640\times480$
    & 64 & 656K & \checkmark & \checkmark & \checkmark \\

    \midrule

    RealDex\cite{liu2024realdex}
    & ROI & 1 & \checkmark & RGB-D & 4 & 52 & --
    & 2.6K & 955K & \checkmark & \checkmark & \checkmark \\

    RoboCOIN\cite{wu2025robocoin}
    & ROI & 15 & \checkmark & RGB-D & 3 & 432 & --
    & 180K & -- & \checkmark & $\times$ & $\times$ \\

    AgiBotWorld\cite{bu2025agibot}
    & ROI & 1 & \checkmark & RGB-D & 8 & 3000 & --
    & 1M & -- & \checkmark & $\times$ & $\times$ \\

    OXE\cite{o2024open}
    & ROI & 22 & $\times$ & RGB-D & 3 & -- & --
    & 1M & 130M & \checkmark & $\times$ & $\times$ \\

    DROID\cite{khazatsky2024droid}
    & ROI & 1 & $\times$ & RGB-D & 3 & -- & $1280\times720$
    & 76K & 56.7M & \checkmark & $\times$ & $\times$ \\

    RoboMIND\cite{wu2025robomind}
    & ROI & 4 & \checkmark & RGB-D & 3 & 96 & $480\times640$
    & 107K & -- & \checkmark & $\times$ & $\times$ \\

    \midrule

    RH20T\cite{fang2024rh20t}
    & HROI & 7 & $\times$ & RGB-D & 7 & -- & $1280\times720$
    & 220K & 50M & \checkmark & $\times$ & $\times$ \\

    DexWild\cite{tao2025dexwild}
    & HROI & 2 & \checkmark & RGB-D & 6 & 180 & $224\times224$
    & 10K & -- & $\times$ & $\times$ & $\times$ \\

    H\&R\cite{xie2026human2robot}
    & HROI & 2 & $\times$ & RGB-D & 1 & -- & $240\times424$
    & 2.6K & 1M & \checkmark & $\times$ & $\times$ \\

    \textbf{HRDexDB (Ours)}
    & \textbf{HROI} & \textbf{5} & \checkmark & \textbf{RGB}
    & \textbf{23} & \textbf{100} & $\mathbf{2048\times1536}$
    & \textbf{3.2K} & \textbf{43.5M} & \checkmark & \checkmark
    & \checkmark \\

    \bottomrule
  \end{tabular}
  }
\end{table*}

\noindent \textbf{Human-Object Interaction Dataset.} 
Human-object interaction datasets have enabled substantial progress in modeling hand articulation and object motion under contact-rich manipulation. Benchmarks such as FreiHAND~\cite{zimmermann2019freihand}, HO-3D~\cite{citahampali2020honnotate}, and DexYCB~\cite{chao2021dexycb} focused on 3D hand-object pose estimation, while other datasets such as ContactPose~\cite{Brahmbhatt_2020_ECCV}, ARCTIC~\cite{fan2023arctic}, HOT3D~\cite{banerjee2025hot3d}, HOI4D~\cite{liu2022hoi4d}, GigaHands~\cite{fu2025gigahands}, TACO~\cite{liu2024taco}, Contact4D~\cite{songcontact4d},  OakInk2~\cite{zhan2024oakink2}, and HO-Cap~\cite{wang2025hocap} expanded the scale, viewpoints, object diversity, contact annotations, and task complexity of hand-object interaction capture. 
These datasets provide rich observations and annotations
for understanding human hand-object interactions.
Complementing this line of work, \textbf{HRDexDB} captures
both human and robotic grasping over shared objects within
a common calibrated setup, supporting the study of how
grasp configurations and contact patterns vary across
embodiments.

\noindent \textbf{Robot-Object Interaction Dataset.}
Robot manipulation datasets have grown substantially in scale and diversity for robot learning.
Large-scale efforts such as Open X-Embodiment~\cite{o2024open} and DROID~\cite{khazatsky2024droid} aggregate diverse demonstrations across many tasks and environments, but much of this data is collected with relatively low-DoF grippers.
More recent datasets such as AgiBot World~\cite{bu2025agibot}, RoboMIND~\cite{wu2025robomind}, and RoboCOIN~\cite{wu2025robocoin} further expand scale and task diversity, supporting bimanual manipulation and demonstrations collected with both grippers and dexterous hands.
Focusing on dexterous grasping, RealDex~\cite{liu2024realdex} collects real-world dexterous robot grasping trajectories through teleoperation, but focuses on a single robotic hand. 
\textbf{HRDexDB} complements these resources by providing detailed 3D annotations of both robotic hand motion and object pose trajectories across multiple dexterous hand embodiments, together with calibrated multi-view RGB observations. These annotations support geometric analysis of robot-object interactions throughout the grasping sequence.

\noindent \textbf{Human-Robot Demonstration Dataset.} 
Recent datasets aim to associate human demonstrations with robotic executions to support learning across embodiments. RH20T~\cite{fang2024rh20t} and H\&R~\cite{xie2026human2robot} provide task- or frame-level human-robot alignment, but are primarily based on parallel-jaw grippers and offer limited coverage of multi-finger dexterous manipulation. 
DexWild~\cite{tao2025dexwild} scales in-the-wild dexterous human demonstrations and aligns human and robot observation and action spaces for co-training, without establishing reference-linked correspondences between individual executions.
 In \textbf{HRDexDB}, human demonstrations guide the collection of robot grasps, while allowing hand-specific configurations and contact patterns. The resulting demonstration-execution associations support downstream studies such as contact map transfer, without assuming identical contacts or frame-level temporal alignment across embodiments.

\noindent \textbf{Synthetic Dexterous Grasp Datasets.}
Large-scale synthetic grasp datasets provide a complementary direction for learning dexterous manipulation priors. 
DexGraspNet~\cite{wang2022dexgraspnet} generates stable dexterous grasps in simulation, while GenDexGrasp/MultiDex~\cite{li2022gendexgrasp} and CEDex~\cite{wu2025cedex} extend grasp synthesis to multiple robotic hand embodiments. 
GraspXL~\cite{zhang2024graspxl} scales grasp motion generation across human and robotic hands.
While synthetic datasets offer substantially greater scale, their generated grasps are not necessarily validated through physical execution. \textbf{HRDexDB} provides recorded real-world trials in which successful grasps are verified by lifting the object with the physical robot, offering empirical evidence of grasp feasibility under the recorded execution conditions.





\section{Constructing HRDexDB}

\subsection{Dataset Overview}
\label{sec:overview}


\textbf{\dataset{}} is constructed as a multi-modal dataset of dexterous grasping sequences performed by both a human subject and four robotic embodiments, spanning \textit{Allegro Hand V5 Plus, Inspire Hand RH56DFTP and RH56F1, }and\textit{ Robotiq 2F-85 gripper}. The dataset contains 43.5M frames and 3.2K sequences over 100 objects, including synchronized visual observations, kinematic states, reconstructed geometry, object 6D poses, and tactile signals when available. All spatial quantities are aligned in a unified world coordinate system defined by our calibrated multi-camera platform.

A human grasping trial is represented as a time-indexed sequence
\begin{equation}
\label{eq:human_data}
\mathcal{T}^{\mathrm{human}} =
\left\{
\{\mat{I}^{c_i}_t\}_{i=1}^{21},\,
\mat{I}^{\mathrm{ego}}_t,\,
\vec{\theta}^{\mathrm{human}}_t,\,
\vec{T}^{\mathrm{object}}_t,\,
y
\right\}_{t=1}^{T_h}.
\end{equation}
Here, $\{\mat{I}^{c_i}_t\}_{i=1}^{21}$ and
$\mat{I}^{\mathrm{ego}}_t$ denote synchronized exocentric
and egocentric RGB observations, respectively.
$\vec{\theta}^{\mathrm{human}}_t \in \mathbb{R}^{51}$
denotes the MANO parameters,
$\vec{T}^{\mathrm{object}}_t \in \SE$ is the object
6D pose, and $y \in \{0,1\}$ indicates grasp success.
$T_h$ denotes the human sequence length.

Similarly, a robotic grasping trial is represented as
\begin{equation}
\label{eq:robot_data}
\mathcal{T}^{\mathrm{robot}} =
\left\{
\{\mat{I}^{c_i}_t\}_{i=1}^{21},\,
\mat{I}^{\mathrm{ego}}_t,\,
\vec{q}^{\mathrm{robot}}_t,\,
\vec{T}^{\mathrm{object}}_t,\,
y
\right\}_{t=1}^{T_r},
\end{equation}
where $\vec{q}^{\mathrm{robot}}_t$ denotes the recorded robot states and commands
and $T_r$ is the robot sequence length. Each robot trial
additionally includes metadata identifying the human
demonstration used as its reference during collection.
This association does not imply frame-level temporal
alignment between the two executions.
Tactile signals $\vec{F}^{\mathrm{tactile}}_t$ are provided
as an additional modality for robots equipped with
fingertip tactile sensors.



\begin{figure}[t]
    \centering
    \includegraphics[width=1.0\linewidth]{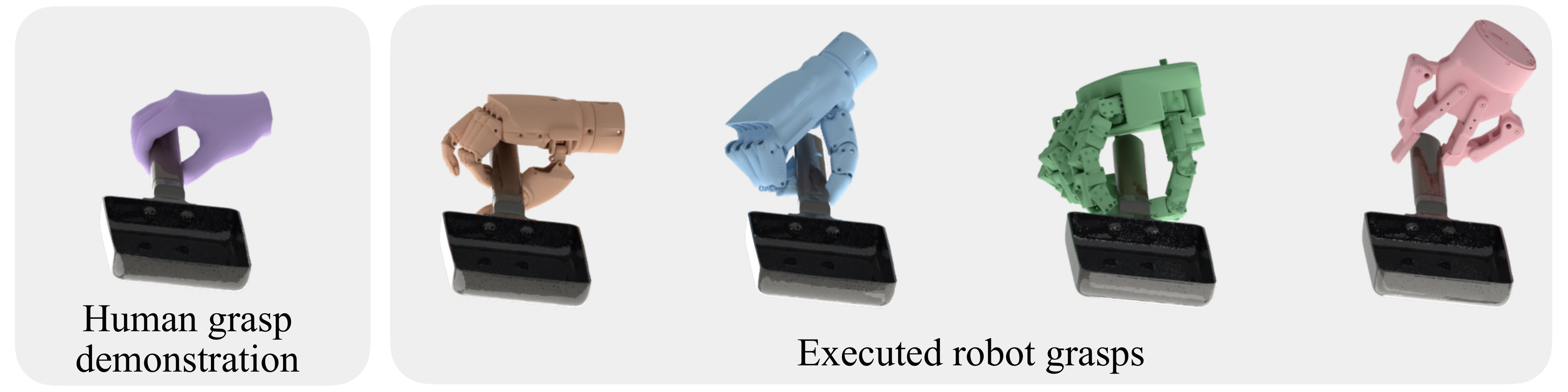}
    \caption{\textbf{Demonstration-guided data acquisition.} A human grasp serves as a reference for teleoperated robot executions, with grasp configurations adapted to each embodiment.}
    \label{fig:placeholder}
    \vspace{-10pt}
\end{figure}
\subsection{Data Acquisition Protocol}
\label{sec:capture_system}

\begin{figure*}[t]
\centering
\includegraphics[width=\linewidth]{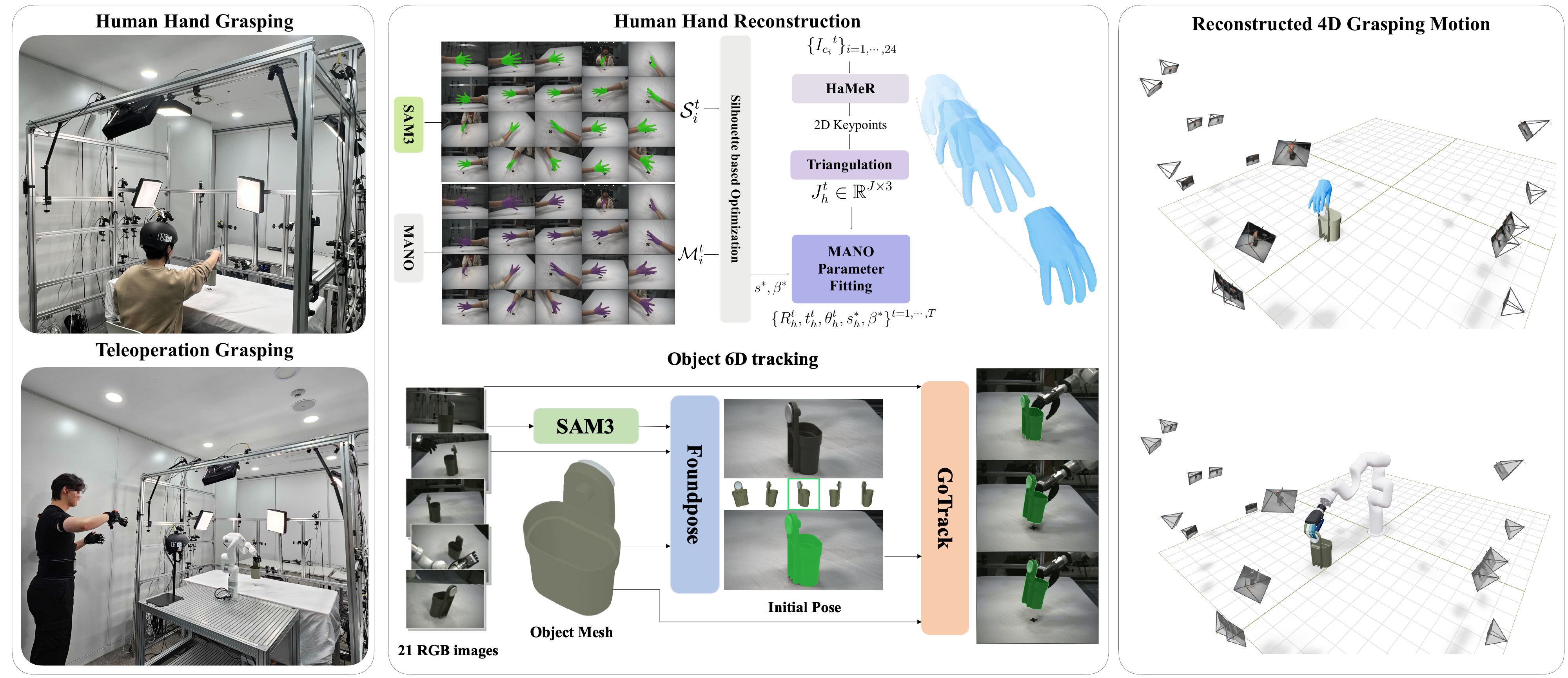}
\captionof{figure}{
    \textbf{Capture and Reconstruction Pipeline}. 
    Multi-view recordings are used to reconstruct hand motion and 6D object trajectories, yielding spatially aligned and temporally synchronized 3D annotations.}
\label{fig:system_overview}
\vspace{-10pt}
\end{figure*}

We collect human and robot grasping trials through a two-stage procedure using the same target object and workspace setup. A human participant first demonstrates a grasp, which is recorded by the multi-view capture system. A teleoperator then observes the demonstration and attempts to reproduce its grasping strategy with the target robotic hand.
Notably, not every human grasp can be reproduced by every robotic hand due to differences in morphology and kinematic constraints. Human demonstrations therefore serve as references, with robot executions allowing embodiment-specific adaptations in finger configuration and contact placement.
Each robot trial is recorded as a separate execution with its own motion trajectory and success/failure outcome. We retain the association between each robot trial and its reference human demonstration as collection metadata.

\subsection{Multi-Modal State Reconstruction}
\label{sec:state_recon}

\noindent\textbf{Capture System.}
Our capture platform (Fig.~\ref{fig:system_overview}) combines a 21-camera RGB rig surrounding the workspace with a stereo helmet for human demonstrations and static stereo egocentric cameras for robotic trials. Camera intrinsics and extrinsics are calibrated using ChArUco observations and COLMAP~\cite{schoenberger2016sfm}, with hand-eye calibration aligning the rig and robot frames. For human sequences, moving egocentric camera poses are recovered through ego-exo feature matching and PnP. Each robotic hand is mounted on an xArm6 manipulator and teleoperated using a Vive wrist tracker and MANUS gloves for wrist and finger control, respectively.

\noindent\textbf{Human Hand Reconstruction.}
To reconstruct 3D human hand motion, we employ the MANO parametric hand model~\cite{MANO:SIGGRAPHASIA:2017}. 
Following the multi-view fitting strategy of GigaHands~\cite{fu2025gigahands}, we detect 2D hand keypoints in each calibrated view using HaMeR~\cite{pavlakos2024reconstructing}, triangulate 3D joints, and optimize MANO pose parameters for each frame. Subject-specific hand shape is calibrated using silhouette alignment with SAM3-generated masks~\cite{carion2025sam3segmentconcepts}, and temporal filtering is applied to reduce jitter.

\noindent\textbf{Robot Hand Reconstruction.} We reconstruct the robot hand geometry through forward kinematics using the recorded joint configurations and the corresponding URDF model. The resulting articulated meshes are transformed into the calibrated world frame using the recorded robot poses.

\noindent\textbf{Object 6D Tracking.}
We estimate object 6D pose trajectories using a model-based multi-view tracking pipeline.
Given multi-view RGB observations, SAM3 \cite{carion2025sam3segmentconcepts} provides object masks at a selected initialization frame to localize the manipulated object. We then estimate its initial 6D pose using FoundPose \cite{ornek2024foundpose} with the object's 3D CAD model, and track the object across the full sequence using GoTrack \cite{nguyen2025gotrack}. At each frame, observations from the selected calibrated views are jointly fused through multi-view triangulation and rigid pose fitting, improving robustness to severe hand-object occlusions and viewpoint-specific tracking failures.

\section{Dataset Analysis}
\begin{figure*}[t]
    \centering
    \includegraphics[width=1.0\linewidth]{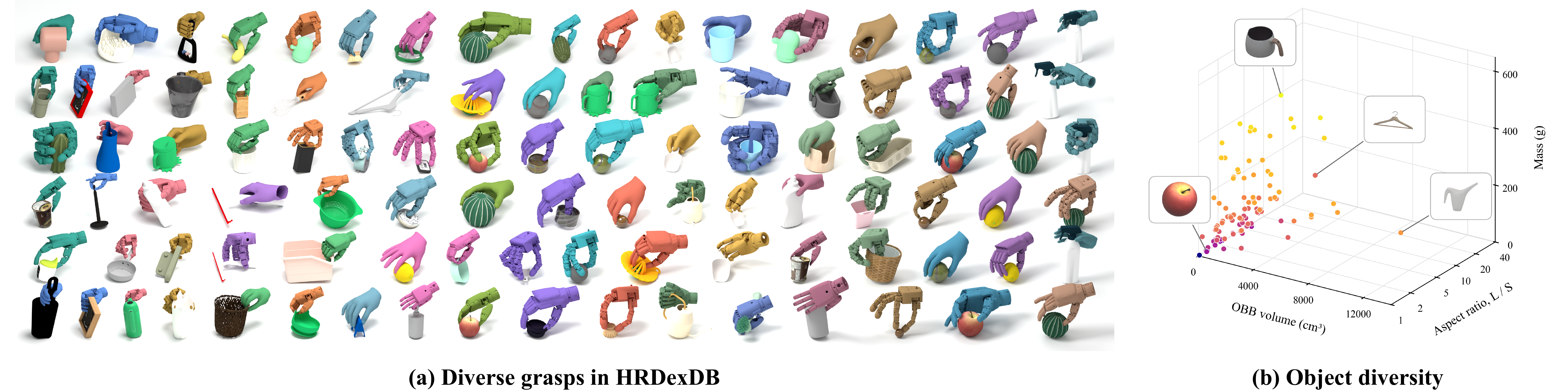}
    \caption{\textbf{Grasp and object diversity in HRDexDB.}
    (a) Human and robotic grasps, colored by grasp type.
    (b) Object distribution by OBB volume, aspect ratio, and mass.
    Aspect ratio is the ratio of the longest to shortest OBB dimension.}
    \label{fig:diversity}
    \vspace{-18pt}
\end{figure*}

\subsection{Dataset Statistics}

\begin{table}[t]
    \centering
    \caption{Trial counts and success rates by embodiment.}
    \label{tab:embodiment_statistics}
    \small
    \begin{tabular}{@{}lrrrr@{}}
    \toprule
    Embodiment & Trials & Successes & Failures & Success (\%) \\
    \midrule
    Human            & 541 & 541 & 0   & 100.00 \\
    Allegro V5       & 539 & 487 & 52  & 90.35 \\
    Inspire RH56F1   & 743 & 513 & 230 & 69.04 \\
    Inspire RH56DFTP & 791 & 476 & 315 & 60.18 \\
    Robotiq 2F-85    & 601 & 474 & 127 & 78.87 \\
    \midrule
    Robot subtotal   & 2,674 & 1,950 & 724 & 72.92 \\
    \textbf{Total}   & \textbf{3,215} & \textbf{2,491}
                     & \textbf{724} & \textbf{77.48} \\
    \bottomrule
\end{tabular}%
    \vspace{-3pt}
\end{table}

HRDexDB comprises 3,215 grasping trials, including 541 human
demonstrations and 2,674 robot executions across four embodiments. 
Table~\ref{tab:embodiment_statistics} summarizes the trial counts
and outcomes for each embodiment. 
All human demonstrations are
successful, while robot trials include both successful and failed
executions.
We observe that the limited thumb mobility of the Inspire hands makes
certain reference pinch grasps difficult to reproduce during
teleoperation, even for experienced teleoperators, which may partly
explain their lower success rates compared with the other embodiments.

\subsection{Grasp and Object Diversity}

HRDexDB covers 23 of the 33 grasp types defined by the GRASP
taxonomy~\cite{feix2015grasp}, including \textit{Precision Sphere}
and \textit{Palmar Pinch}.
Figure~\ref{fig:diversity}(a) illustrates this diversity through
representative human and robotic grasps.
We will release these grasp-type labels with the dataset.

The dataset includes 100 everyday objects spanning kitchenware,
household items, toys, tools, and organic objects.
Figure~\ref{fig:diversity}(b) shows their distribution in OBB volume,
aspect ratio, and mass.
OBB volumes range from 0.11 to 8.70~L (median 1.31~L),
while masses range from 3 to 533~g (median 107~g).

\subsection{Object 6D Pose Tracking Quality}
\begin{table}[t]
    \centering
    \caption{Object silhouette consistency (mean IoU $\uparrow$)
    against manually verified SAM3 masks.}
    \label{tab:silhouette_consistency}
    \small
    \setlength{\tabcolsep}{7pt}
    \begin{tabular}{@{}lcc@{}}
        \toprule
        Embodiment & FoundationPose & Ours \\
        \midrule
        Human            & 0.841 & \textbf{0.915} \\
        Allegro V5       & 0.742 & \textbf{0.905} \\
        Inspire RH56F1   & 0.894 & \textbf{0.942} \\
        Inspire RH56DFTP & 0.723 & \textbf{0.908} \\
        Robotiq 2F-85    & 0.805 & \textbf{0.933} \\
        \midrule
        \textbf{Overall} & 0.801 & \textbf{0.921} \\
        \bottomrule
    \end{tabular}
    \vspace{-18pt}
\end{table}
We evaluate our 6D object pose tracking by measuring image-space consistency on 1,500 images, 300 per embodiment (Table~\ref{tab:silhouette_consistency}).
We compare our multi-view GoTrack pipeline with a two-view FoundationPose~\cite{wen2024foundationpose} tracking baseline, using identical evaluation
images and occlusion handling.
We compute IoU between depth-aware rendered object silhouettes
and manually verified SAM3 masks in the capture views used
for annotation.
Our pipeline achieves a mean IoU of 0.921, compared with
0.801 for FoundationPose, with improvements across all five embodiments.





\section{Experiments \& Benchmarking}
\label{sec:applications}

We evaluate HRDexDB through three downstream tasks: human-to-robot contact map transfer, visual robot-object contact estimation, and retrieval-assisted grasping. 

\subsection{Human-to-Robot Contact Map Transfer}

\begin{figure*}[t]
\centering
  \includegraphics[width=\linewidth]{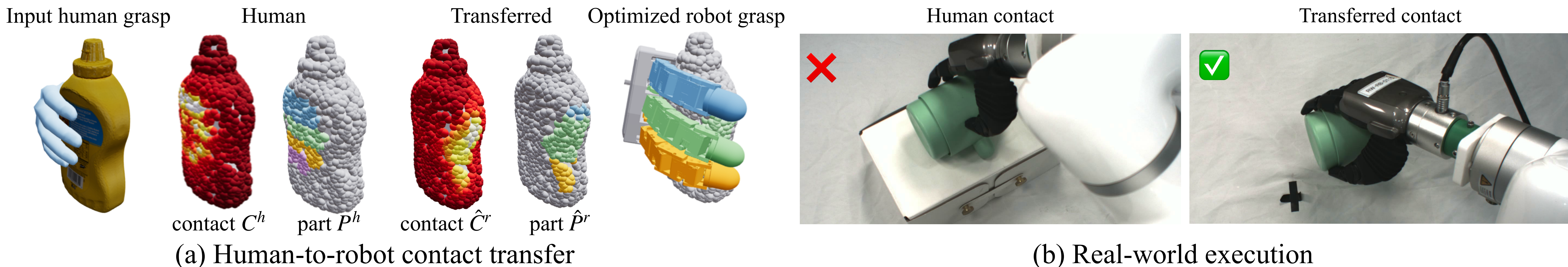}
  \caption{\textbf{Human-to-robot contact transfer and real-world grasping.}
(a)~Human contact and part maps $(C^h, P^h)$ are transferred to
robot-specific maps $(\hat{C}^r, \hat{P}^r)$ for grasp optimization.
(b)~For the same object and human grasp, transferred contacts yield
a successful grasp, whereas direct human contacts fail.}
  \label{fig:transfer_overview}
  \vspace{-16pt}
\end{figure*}

\noindent\textbf{Task definition.}
Human contact patterns provide useful references for robot
grasping, but differences in hand morphology and kinematics
can make direct imitation suboptimal.
We investigate whether HRDexDB enables learning
robot-specific contact representations from human grasps.
Given an object and an observed human contact representation,
the task is to predict contact and hand-part maps for a
target robotic hand, which subsequently guide grasp synthesis.

\noindent\textbf{Experimental setup.}
We use the demonstration-execution associations in HRDexDB
to supervise human-to-robot contact map transfer.
The associated robot contact map serves as the target for each reference human grasp, allowing
embodiment-specific differences in contact placement and finger configuration.

We represent a grasp on an object point cloud $O \in \mathbb{R}^{N \times 3}$ with a contact map $C \in [0,1]^{N}$ of per-point contact probabilities and a part map $P$ assigning contacted points to hand parts. The human part map is $P^h \in \mathbb{R}^{N \times 6}$, while the robot part map is $P^r \in \mathbb{R}^{N \times B}$, with $B=6$ for the Inspire hand and $B=5$ for the Allegro hand. Conditioned on the human representation $[C^h, P^h]$ and PointNet++~\cite{qi2017pointnetdeephierarchicalfeature} object features, our model predicts the robot representation $[C^r, P^r]$, supervised with a contact-weighted $L_1$ loss on $C^r$ and a cross-entropy loss on $P^r$ over contacted points.
 We train a separate model for each target hand, using the Inspire RH56F1 and Allegro Hand V5 Plus.

Given the predicted robot contact representation, we synthesize grasps using the physics-aware optimization pipeline of CEDex~\cite{wu2025cedex}, which combines contact, penetration, and self-collision terms. We compare \emph{Human-Contact}, which optimizes against the captured human contact map, with \emph{Transferred-Contact}, which optimizes against our predicted robot-specific contact map. The optimizer is fixed across both settings, isolating the effect of the contact objective. For the four-fingered Allegro hand, the baseline maps the fourth finger to the union of the human ring and little fingers.

We evaluate real-world grasp success using held-out human grasps
of objects seen during training as inputs.
A grasp succeeds if the object is lifted and held for ten seconds. We generate pre-grasp and squeeze motions with BODex~\cite{chen2025bodex} and compute execution trajectories with CuRobo~\cite{curobo_v2}.

\begin{table}[t]
  \centering
  \caption{Real-world grasp success rates (\%) for human-to-robot contact map transfer, evaluated over 60 trials per hand and condition. Human and transferred contact maps guide the same grasp optimizer.}
  \label{tab:contact_transfer_success}
  \small
  \setlength{\tabcolsep}{4pt}
  \begin{tabular}{lcc}
\toprule
\textbf{Method}
  & \textbf{Inspire} $\uparrow$
  & \textbf{Allegro} $\uparrow$ \\
\midrule
Human-Contact
  & $66.7$ & $60.0$ \\
Transferred (Ours)
  & $\mathbf{73.3}$ & $\mathbf{70.0}$ \\
  \midrule
Improvement        & +6.7 & +10.0 \\
\bottomrule
\end{tabular}

  \vspace{-3pt}
\end{table}

\begin{table}[t]
    \centering
    \caption{Contact transfer generalization. Grasp success rate (\%) on unseen objects. }
    \label{tab:unseen_object}
    \small
    \setlength{\tabcolsep}{4pt}
    \begin{tabular}{lcc}
\toprule
\textbf{Method} & \textbf{DexYCB (20 objs.)} & \textbf{GRAB (48 objs.)} \\
\midrule
Human-Contact      & 53.69 & 50.74 \\
Transferred (Ours) & \textbf{55.94} & \textbf{56.81} \\
\midrule
Improvement        & +2.25 & +6.07 \\
\bottomrule
\end{tabular}

    \vspace{-12pt}
\end{table}


\noindent\textbf{Results.}
Figure~\ref{fig:transfer_overview} illustrates how transferred
contact maps adapt the human contact distribution to the target robot hand.
Table~\ref{tab:contact_transfer_success} summarizes grasp success rates under the two contact objectives. Transferred-Contact improves success over directly using human contact maps, showing that HRDexDB enables learning robot-specific contact strategies from demonstration guided robot grasps.

We further evaluate contact transfer with Allegro V5 on 20 unseen
objects from DexYCB~\cite{chao2021dexycb} and 48 from GRAB~\cite{taheri2020grab} in simulation.
For each object, we use 10 distinct human grasps as inputs and
generate a robot grasp with each of eight random seeds per input,
yielding 80 robot grasps per object.
Following CEDex, we apply forces along the positive and negative directions of the three coordinate axes in Isaac Gym~\cite{makoviychuk2021isaac}.
Transferred contact outperforms human contact on both datasets (Table~\ref{tab:unseen_object}), demonstrating that contact transfer learned from HRDexDB generalizes to unseen objects.



\subsection{Visual Robot-Object Contact Estimation}

\noindent\textbf{Task definition.}
Tactile sensing is an essential component of dexterous
manipulation~\cite{lee2024dextouch, mack2025visuo, miller2025enhancing},
yet tactile cues are largely absent from existing robot manipulation methods because reliable, readily available sensors are scarce. Recently, vision-based contact and pressure estimation has been explored for human hands~\cite{jeon2026hope}. We adopt an analogous approach for robot hands and predict binary contact directly from visual observations, which can potentially reduce the need to design, mount, and calibrate tactile sensors for each robot hand. We benchmark binary contact prediction for 60 hand
regions from a single RGB frame on the Inspire RH56F1 and Allegro V5 hands, using hand-specific partitions that are finer near the fingertips and distal links.

\noindent\textbf{Contact annotation.}
\footnote{The Inspire and Allegro hands are equipped with tactile sensors, but their readings are too noisy or sparse to serve as reliable contact labels. We still release these signals with HRDexDB.}
\begin{figure}[t]
    \centering
    \includegraphics[width=1.0\linewidth]{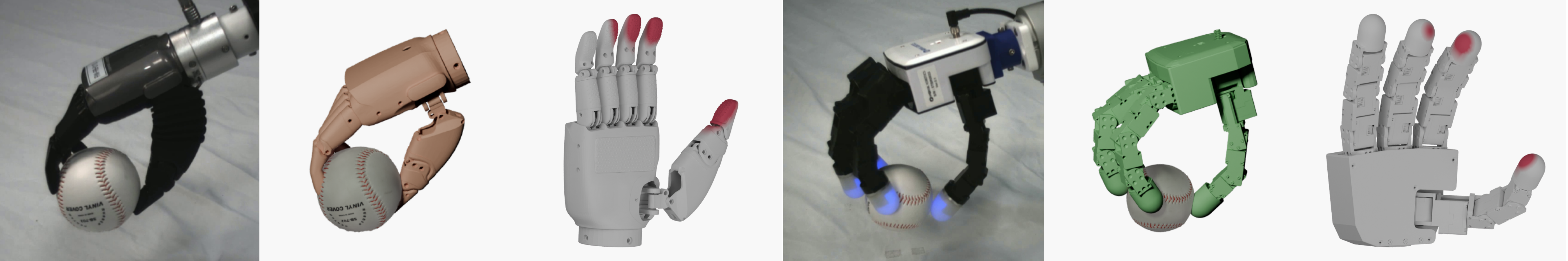}
    \caption{\textbf{Geometry-derived contact annotation.} From left to right: RGB observations, reconstructed hand-object geometry, and region-level contact labels for Inspire and Allegro hands. Regions within 3~mm of the object surface are labeled as contacting and highlighted in red.}
    \label{fig:contact_annotation}
    \vspace{-12pt}
\end{figure}
As illustrated in Figure~\ref{fig:contact_annotation}, we derive
region-level contact labels from the reconstructed 4D hand-object
geometry. A hand region is labeled as contacting when its distance
to the object surface is at most 3~mm. This provides supervision
for both visible and occluded hand regions in the input RGB image.

\noindent\textbf{Baselines and experimental setup.}
We adapt HOPE~\cite{jeon2026hope}, HACO~\cite{jung2025haco}, and PressureVision~\cite{grady2022pressurevision} to predict 60 region-level contact labels. 
For HOPE, we replace the human-hand vertex tokens with 60 robot surface-region tokens conditioned on their articulated palm frame positions and the camera-relative palm pose, removing temporal attention and pressure prediction for single-frame binary contact estimation. 
For HACO, we replace the MANO-based~\cite{MANO:SIGGRAPHASIA:2017} vertex outputs and multi-resolution contact mappings with 60 robot surface-region outputs. 
PressureVision uses robot state to project the hand mesh for dense supervision and region-level prediction. At inference, we average logits sampled at each region's projected vertices, treating samples outside the image or behind the camera as non-contact.

Inspire RH56F1 includes 74/18 objects and 56.1K/14.6K images for training/validation; Allegro V5 includes 72/19 objects and 37.3K/8.2K images.
HOPE, HACO, and PressureVision are trained for 20, 10, and 60 epochs, respectively.
We report micro-averaged precision, recall, and F1 over all valid validation frame-region pairs.

\noindent\textbf{Results.}
\begin{table}[t]
  \caption{Region-level contact estimation on two robot-hand embodiments (Inspire RH56F1/Allegro V5).}
  \label{tab:robot_contact_results}
  \centering
  \setlength{\tabcolsep}{4.2pt}
  \renewcommand{\arraystretch}{1.10}
  \resizebox{\columnwidth}{!}{%
    \begin{tabular}{lllccc}
      \hline
      Embodiment & Method & Input
      & F1 & Precision & Recall \\
      \hline
      \multirow{3}{*}{Inspire RH56F1}
      & PressureVision & RGB + robot state & 0.594 & \textbf{0.791} & 0.476 \\
      & HACO           & RGB         & 0.414 & 0.278 & \textbf{0.810} \\
      & HOPE           & RGB + robot state & \textbf{0.679} & 0.689 & 0.669 \\
      \hline
      \multirow{3}{*}{Allegro V5}
      & PressureVision & RGB + robot state & 0.424 & \textbf{0.707} & 0.303 \\
      & HACO           & RGB         & 0.476 & 0.338 & \textbf{0.806} \\
      & HOPE           & RGB + robot state & \textbf{0.644} & 0.681 & 0.612 \\
      \hline
    \end{tabular}%
  }
  \vspace{-1mm}
\end{table}

HACO achieves high recall but low precision, whereas
PressureVision shows the opposite trade-off.
HOPE achieves the highest F1 score on both embodiments,
with a balanced precision-recall trade-off.
These comparisons reflect the adapted pipelines,
including differences in auxiliary inputs and region-level
prediction procedures.
Overall, the results establish baselines for contact estimation
on unseen objects and demonstrate how HRDexDB's geometric
annotations enable supervision across robotic-hand embodiments.





\subsection{Retrieval-Assisted Grasping}
\begin{figure}[t]
    \centering
    \includegraphics[width=1.0\linewidth]{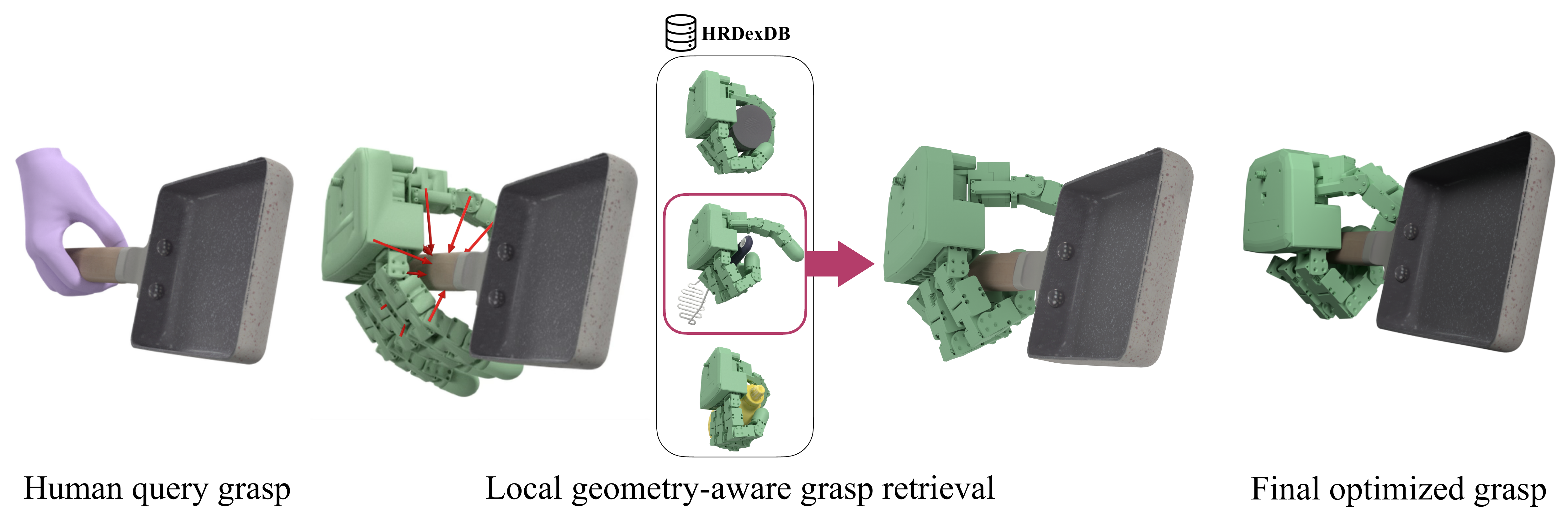}
    \caption{\textbf{Retrieval-Assisted Grasping.} Given a human grasp, we retrieve robot grasps from HRDexDB using local object geometry, align them to the target object, and refine hand-object contacts to generate a robot grasp.}
    \label{fig:rag}
    \vspace{-18pt}
\end{figure}
\noindent\textbf{Task definition.}
Human grasps provide a natural prior for where and how a robot should grasp an object. However, kinematic retargeting of human grasps may fail to establish stable contacts due to differences in hand morphology and kinematics. We investigate whether robot grasps recorded in HRDexDB can complement retargeted grasps with morphology-compatible priors. Given a human grasp and the target object mesh, we formulate \emph{Retrieval-Assisted Grasping (RAG)}, which generates a robot grasp near the demonstrated one by retrieving and adapting database grasps with similar local object geometry.



\noindent\textbf{Method.}
RAG combines geometric retrieval, local alignment, and contact refinement (Fig.~\ref{fig:rag}).
We first obtain a robot wrist pose $\mathbf{T}_q$ from the
human grasp using AnyTeleop~\cite{qin2023anyteleop}.
Inspired by hand-centric object representation of RobustDexGrasp~\cite{zhang2025RobustDexGrasp}, 
we construct a local geometry descriptor for grasp retrieval. 
We place the robot hand at $\mathbf{T}_q$ with a fixed canonical finger configuration $\mathbf{q}_{\mathrm{can}}$, sample $N_h=168$ hand surface points, and sample $N_o=4096$ area-weighted points from the object mesh.
For each hand point $\mathbf{x}_i$, we find the nearest
sampled object point $\mathbf{y}_i$ and construct a descriptor
from the relative vectors $\mathbf{y}_i-\mathbf{x}_i$
and their Euclidean distances.

We retrieve the top 16 database grasps and process them
in ranked order.
For each candidate, we extract object patches within
4~cm of the hand surface and align the database patch
to the query patch using FPFH~\cite{5152473} matching,
RANSAC, and point-to-plane ICP.
We transform the database wrist pose using the estimated
alignment while retaining its finger configuration.
We retain up to six candidates whose aligned wrist poses
lie within 2~cm and $15^\circ$ of $\mathbf{T}_q$.

For each retained candidate, we identify contact anchors
as hand surface points within 5~mm of the database object. We jointly refine the wrist pose and finger configuration to bring these anchors into contact with the query object, while penalizing object penetration and hand self-collision.
The optimized candidate with the most hand-object contacts is selected for execution.

\noindent\textbf{Experimental setup.}
To isolate the contribution of retrieval, we compare our \emph{RAG} against \emph{Kinematic Retargeting + Optimization}, which applies the same contact-aware optimization used in RAG directly to the kinematically retargeted grasp, without database retrieval. 
We use the Allegro V5 hand for all experiments. For evaluation, we split the 100 objects in HRDexDB into 80 database objects and 20 unseen test objects. The database objects are used only for grasp retrieval, while evaluation is restricted to the 20 unseen objects. In simulation, we evaluate five human grasps per test object in MuJoCo~\cite{6386109}, resulting in 100 trials in total. For real-world evaluation, we use one human grasp per object, resulting in 20 trials in total. In both settings, a trial is considered successful if the robot successfully grasps and lifts the object.

\noindent\textbf{Results.}
\begin{table}[t]
    \centering
    \caption{Grasping success rate on 20 unseen objects.}
    \label{tab:retrieval_assisted_grasping}
    \begin{tabular*}{\linewidth}{@{\extracolsep{\fill}}lcc}
        \toprule
        \textbf{Method}
          & \multicolumn{2}{c}{\textbf{Success Rate} $\uparrow$} \\
        \cmidrule(lr){2-3}
          & \textbf{Sim.} & \textbf{Real} \\
        \midrule
        Retargeting + Opt.
          & $0.50$ & $0.40$ \\
        RAG (Ours)
          & $\mathbf{0.73}$ & $\mathbf{0.75}$ \\
        \midrule
        Improvement
          & $+0.23$ & $+0.35$ \\
        \bottomrule
    \end{tabular*}
    \vspace{-18pt}
\end{table}
Table~\ref{tab:retrieval_assisted_grasping} shows that RAG
improves simulated grasp success from 50/100 (50\%) to
73/100 (73\%) and real-world success from 8/20 (40\%)
to 15/20 (75\%).
These results support the utility of robot grasp priors retrieved from HRDexDB for grasp synthesis on unseen objects, beyond refining kinematically retargeted grasps alone, and highlight the potential of retrieval-based priors for improving grasp generalization to novel object geometries.

\section{Conclusion}
\label{sec:conclusions}

We present \textbf{HRDexDB}, a real-world 4D dataset of human and robotic dexterous grasping across multiple embodiments, comprising 3.2K sequences over 100 objects with detailed 3D hand and 6D object annotations. 
By capturing diverse hands interacting with shared objects, HRDexDB supports the study of embodiment-dependent grasp strategies and contact patterns. 
Our experiments demonstrate its utility for cross-embodiment contact map transfer, visual contact estimation, and retrieval-assisted grasping.
The dataset and benchmark protocols will be publicly released.
Moving forward, we plan to expand HRDexDB toward 1,000 objects and more complex functional manipulation tasks.









\bibliographystyle{IEEEtran}
\bibliography{main}

\end{document}